\documentclass[letterpaper, 10 pt, conference]{ieeeconf}  

\IEEEoverridecommandlockouts                              

\usepackage{graphics} 
\usepackage{epsfig} 
\usepackage{mathptmx} 
\usepackage{times} 
\usepackage{amsmath} 
\usepackage{amssymb}  
\usepackage{amsfonts}
\usepackage{threeparttable}
\usepackage{textcomp}
\usepackage{siunitx}
\usepackage{adjustbox}
\usepackage{diagbox}
\usepackage{booktabs, multirow}
\usepackage{adjustbox}
\usepackage{booktabs, multirow}
\usepackage{xcolor}
\usepackage[utf8]{inputenc}
\usepackage{graphicx}
\usepackage{tikz}
\usepackage{listings}                       
\usepackage{algorithm, algorithmic}                      
\usetikzlibrary{backgrounds}

\usepackage[noadjust]{cite}
\usepackage{gensymb}

\title{\LARGE \bf
DA-Fusion: Deformable Attention-based RGB-D Fusion Transformer for Unseen Object Instance Segmentation}
\author{Yesol Park$^{1*}$ Hye-Jung Yoon$^{1*}$ Juno Kim$^{1*} $ Byoung-Tak Zhang$^{1,2,3}$
\thanks{This work was partly supported by the IITP (RS-2021-II212068-AIHub/10\%, RS-2021-II211343-GSAI/15\%, RS-2022-II220951-LBA/15\%, RS-2022-II220953-PICA/20\%), NRF (RS-2024-00353991-SPARC/20\%, RS-2023-00274280-HEI/10\%), and KEIT (RS-2024-00423940/10\%) grant funded by the Korean government.
}
\thanks{*Authors have equal contributions}
\thanks{$^{1}$Interdisciplinary Program in AI, Seoul National University}%
\thanks{$^{2}$Artificial Intelligence Institute, Seoul National University}%
\thanks{$^{3}$Department of Computer Science, Seoul National University}%
}
\begin{document}

\maketitle
\thispagestyle{empty}
\pagestyle{empty}

\begin{abstract}
In logistics automation, precise segmentation of unseen objects is crucial for efficient robotic manipulation in cluttered environments. Tasks such as bin-picking and shelf-picking require robust perception to handle occlusions, varying object shapes, and complex spatial arrangements. Traditional RGB-based methods tend to over-segment objects due to their reliance on texture, while depth-based methods often under-segment by focusing primarily on geometric features. To address these limitations, we propose DA-Fusion, a deformable attention-based RGB-D fusion Transformer designed for unseen object instance segmentation. DA-Fusion effectively combines the strengths of both RGB and depth data, enhancing segmentation accuracy in cluttered and multi-layered object environments. We also introduce the Object Clutter Bin Dataset (OCBD), a benchmark dataset specifically tailored for evaluating bin-picking scenarios in top-down views. Extensive evaluations demonstrate that DA-Fusion outperforms state-of-the-art methods across diverse environments, making it particularly suited for real-world logistics tasks.
\end{abstract}

\section{INTRODUCTION}
Accurately segmenting unseen objects is critical for enabling robots to perform tasks like bin-picking, shelf-picking, and warehouse sorting in logistics. These operations often occur in cluttered, dynamic environments where objects may be occluded, stacked, or vary in shape and texture. The ability to reliably identify and manipulate unseen objects in such conditions is essential for efficient logistics but remains challenging for current robotic systems.

Traditional segmentation methods typically rely on either RGB or depth data, each of which presents its own limitations. RGB-based methods are prone to over-segmentation due to texture sensitivity, especially in environments with complex visual details. In contrast, depth-based methods tend to under-segment by emphasizing geometric features, particularly when objects have similar shapes or are placed closely together. Combining these two modalities to take advantage of their complementary strengths is a key challenge in object segmentation~\cite{wang2018depth, wang2021brief}.

To address this challenge, we propose DA-Fusion, a novel deformable attention-based RGB-D fusion Transformer designed for unseen object instance segmentation (UOIS). DA-Fusion employs a deformable attention mechanism~\cite{zhu2020deformable} to dynamically fuse RGB and depth data at multiple levels within the model. This approach enables more precise and fine-grained integration of texture and geometric information, significantly enhancing segmentation accuracy in the complex environments typical of logistics operations.

\begin{figure} [t!]
\centering
\includegraphics[width=0.8\columnwidth]{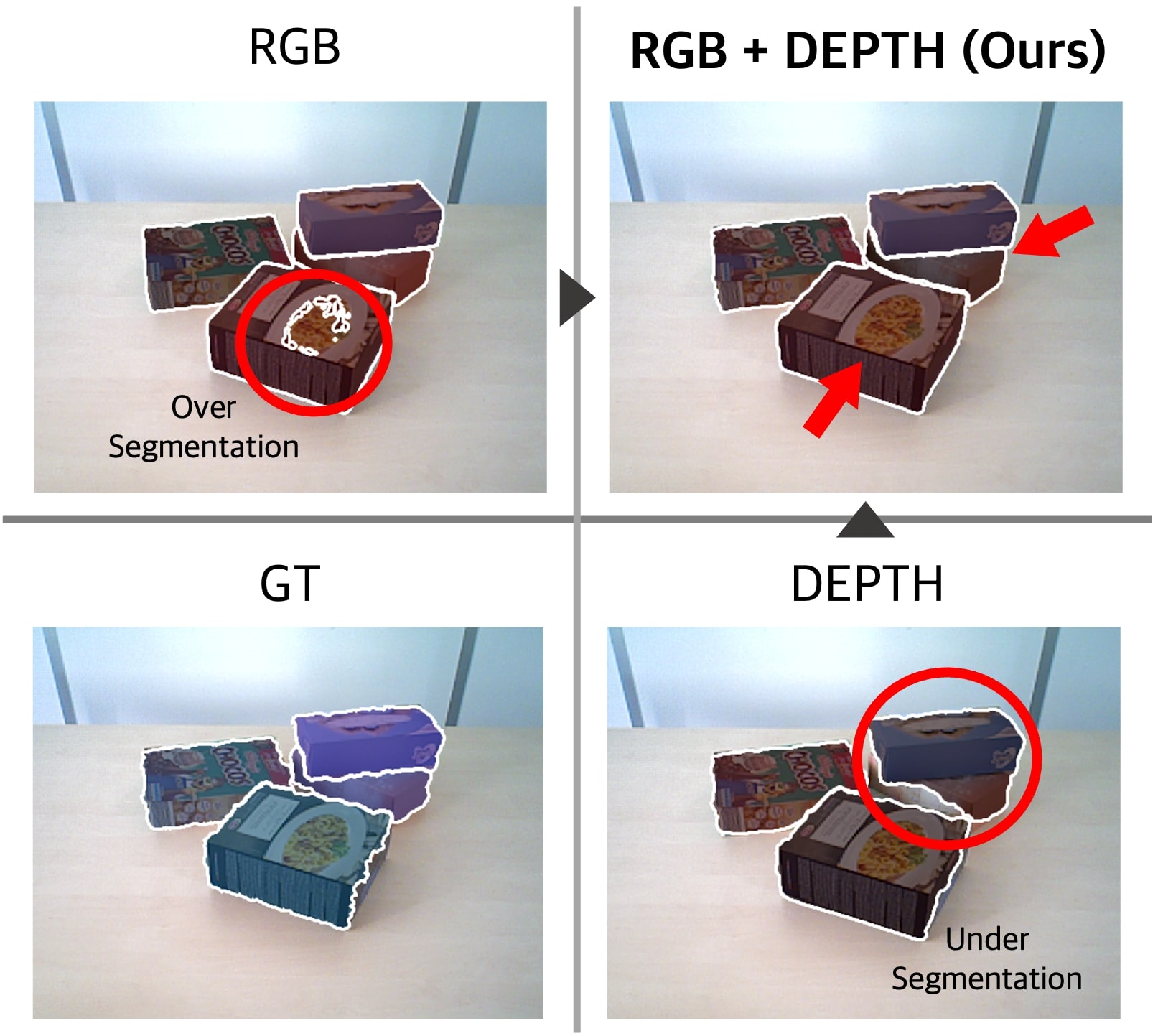}
\vspace{-2mm}
\caption{\textbf{Unseen Object Segmentation results on~\cite{suchi2019easylabel}.} RGB-based segmentation tends to over-segment (top left), while depth-based under-segments (bottom right). Our fusion strategy improves accuracy (top right). Note that GT (bottom left) is auto-labeled~\cite{suchi2019easylabel}.}
\label{problem_setting}
\vspace{-3mm}
\end{figure}

As illustrated in Fig.~\ref{problem_setting}, DA-Fusion provides significant improvements in segmentation performance over using RGB or depth data alone. By dynamically fusing these modalities using deformable attention, DA-Fusion aligns more closely with the Ground Truth (GT), particularly in environments where objects display varied textures and geometries.
 
Additionally, we introduce the Object Clutter Bin Dataset (OCBD), a new real-world dataset designed to evaluate segmentation performance in challenging bin-picking scenarios. Unlike previous datasets~\cite{xu2022fpcc, xu2020convolutional, chen2021metagraspnet_v0, gilles2022metagraspnet}, which are mainly composed of simple, single-type objects with uniform colors and are often created in simulated environments, OCBD captures real-world complexity, including complex and diverse objects with varied colors, multi-layered object arrangements, and noisy RGB-D data, making it a valuable benchmark for real-world applications.

DA-Fusion sets a new state-of-the-art in UOIS across tabletop~\cite{richtsfeld2012segmentation}, indoor~\cite{suchi2019easylabel}, and bin environments, showcasing its effectiveness and versatility in a diverse range of settings. It excels in front-view scenarios, such as tabletop and shelf-picking tasks, as well as in top-down views commonly used in bin-picking operations, further demonstrating its adaptability to various robotic applications.

Our key contributions are summarized as follows:
\begin{itemize}
    \item We propose DA-Fusion, a deformable attention-based Transformer that fuses RGB and depth for improved UOIS accuracy.
    \item We introduce OCBD, a benchmark with top-down views of cluttered, multi-layered objects for real-world bin-picking evaluation.
    \item We validate DA-Fusion on three benchmarks—tabletop, indoor, and bin—achieving state-of-the-art performance.
\end{itemize}

\section{RELATED WORKS}
 
\subsection{Unseen Object Instance Segmentation (UOIS)}
Unseen Object Instance Segmentation (UOIS) is crucial for robotic systems in diverse environments such as tables, shelves, and bins~\cite{zhang2016dorapicker, causo2018robust, yoon2024seg2grasp, li2022sim}. UOIS segments objects without prior knowledge of their classes or appearances, enabling robots to handle previously unseen objects in dynamic settings.

Earlier UOIS methods extended Mask R-CNN~\cite{he2017mask} to generalize beyond trained categories but struggled with small objects and densely packed scenes~\cite{xie2020best, xie2021unseen, back2022unseen}. Transformer-based architectures like Mask2Former~\cite{cheng2021maskformer, cheng2022masked} improved this by better handling complex object layouts through multi-scale feature extraction.

Some recent UOIS methods incorporated second-stage refinement techniques~\cite{xiang2021learning, lu2022mean} to improve accuracy, but this added computational complexity, especially in cluttered scenes. Despite these improvements, reliable segmentation of unseen objects remains challenging, particularly in scenarios with overlapping or ambiguous boundaries.

Our approach builds on the Mask Transformer~\cite{cheng2022masked}, enhancing its ability to handle complex layouts and boundaries, making it suitable for challenging, real-world applications.

\subsection{RGB-D Fusion}
RGB-D fusion is an essential technique for improving segmentation by combining the complementary strengths of RGB and depth data. RGB images provide detailed color and texture information, while depth contributes spatial and geometric insights. Traditional RGB-D fusion approaches are typically divided into early, intermediate, and late fusion strategies.

Early fusion combines RGB and depth data at the input stage, allowing for joint feature extraction, but often fails to fully exploit the distinct advantages of each modality~\cite{sun2020fuseseg}. Intermediate fusion merges RGB and depth information at various stages within the network, enabling better interaction between low-level and high-level features, but increases computational complexity~\cite{zhang2023cmx}. Late fusion processes RGB and depth streams separately and merges them only at the final output. While this method retains modality-specific characteristics, it lacks rich cross-modal interactions, which can limit performance in complex environments~\cite{tziafas2023early}.

Attention mechanisms have been introduced in multi-modal fusion tasks to improve how models integrate RGB and depth data, allowing them to dynamically focus on the most important regions across modalities. While traditional Vision Transformers (ViTs)~\cite{dosovitskiy2020image} primarily use self-attention within a single modality, cross-attention has been employed in tasks involving long-range dependencies between different modalities, particularly in RGB-D fusion and other multi-modal scenarios~\cite{zhao2023cross, zhang2023cmx}. However, attention mechanisms can be computationally expensive, especially in cluttered environments like logistics scenarios.

Deformable attention, introduced in Deformable DETR~\cite{zhu2020deformable}, addresses these limitations by focusing on a small set of key sampling points, reducing computational costs while allowing the model to focus on spatially important regions. This mechanism proves particularly effective in cluttered or occluded environments, where it captures finer-grained details and object parts that are difficult to discern, enhancing the model’s ability to segment small or partially hidden objects.

In our work, DA-Fusion, we fuse the complementary strengths of RGB and depth data using deformable attention. By dynamically focusing on key spatial regions in both modalities, deformable attention allows the model to better capture the rich texture and color from RGB and the spatial and structural insights from depth. This results in more precise fusion, enhancing segmentation accuracy for both large objects and smaller, occluded parts, particularly in cluttered environments, while maintaining computational efficiency.

\begin{figure*} [t!]
\centering
\includegraphics[width=0.97\textwidth]{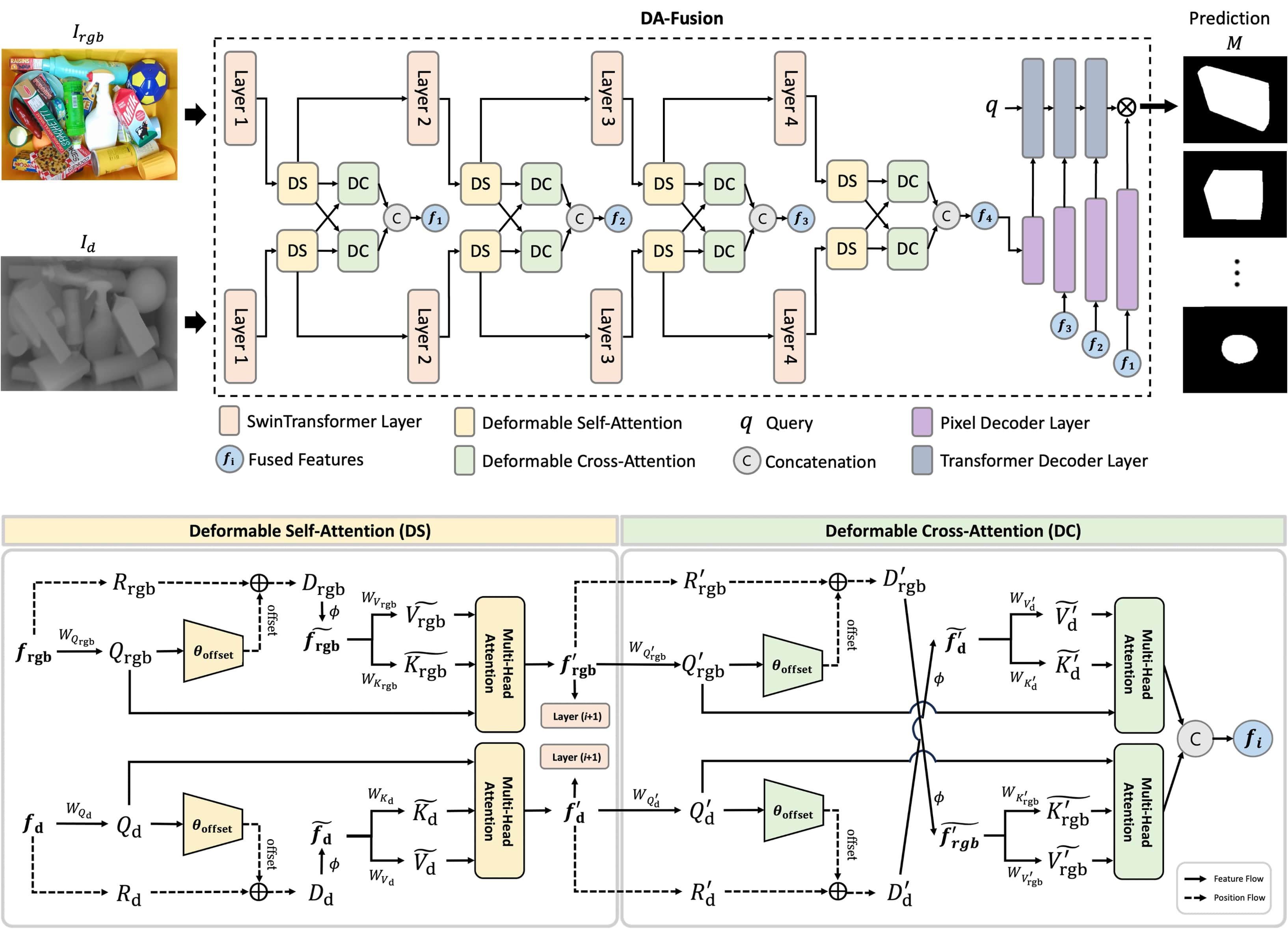}
\vspace{-2mm}
\caption{\textbf{Overview of the DA-Fusion Architecture.} Two parallel branches extract features from RGB and depth inputs, which are fused at multiple layers using deformable attention. The fused features are processed by the Mask Transformer decoder~\cite{cheng2022masked} to generate segmentation masks.}
\label{structure}
\vspace{-4mm}
\end{figure*}

\section{METHOD}
We propose DA-Fusion, a novel framework for RGB-D fusion aimed at unseen object instance segmentation. Our approach utilizes deformable attention mechanisms to refine and fuse features from RGB and depth, capturing complementary information across layers in the encoder. DA-Fusion enables robust and discriminative feature representations by effectively capturing both local and global interactions.

\subsection{Notation Overview}
Let \( {I}_{\text{rgb}} \in \mathbb{R}^{H \times W \times 3} \) and \( {I}_{\text{d}} \in \mathbb{R}^{H \times W \times 3} \) represent the input RGB and depth images, with \( {I}_{\text{d}} \) in HHA format~\cite{gupta2014learning}. Feature maps from the 4-layer backbone are denoted as \( {f}_{\text{rgb}} \) and \( {f}_{\text{d}} \) with dimensions \( H \times W \times C \), where \( C \) varies by layer. For stage $i\!\in\!\{1..4\}$, $f^i_{\text{rgb}},f^i_{\text{d}}\in\mathbb{R}^{H_i\times W_i\times C_i}$ and we denote fused features as $f_i$ accordingly.

The Deformable Self-Attention (DS) module refines these features into \({f}_{\text{rgb}}' \) and \( {f}_{\text{d}}' \), which are processed by the Deformable Cross-Attention (DC) module to fuse information across modalities. The fused features at each layer are denoted as \( {f}_i \) for \( i \in \{1, 2, 3, 4\} \), which are passed into the decoder to produce the final masks \( {M} \).

\subsection{Framework Overview}
Fig.~\ref{structure} shows the DA-Fusion framework, where two parallel branches extract features from \( {I}_{\text{rgb}} \) and \( {I}_{\text{d}} \). The backbone is a 4-layer Swin Transformer~\cite{liu2021swin} that processes features through the DS and DC modules, which fuse RGB and depth information. Outputs from all four layers, \( f_1, f_2, f_3, f_4 \), are concatenated and decoded by the pixel decoder of Mask Transformer~\cite{cheng2022masked}, with the final mask \( {M} \) generated by the transformer decoder.

\subsection{Deformable Self-Attention (DS)}
In Deformable Self-Attention (DS), the feature maps \( {f}_{\text{rgb}} \) and \( {f}_{\text{d}} \) are refined through attention mechanisms that dynamically adjust the sampling points based on learned offsets. Specifically, the refined features, \( {f}_{\text{rgb}}' \) and \( {f}_{\text{d}}' \), are generated by applying deformable attention to adapt the key and value embeddings at deformed positions.

The reference points $R_{\text{rgb}}$ and $R_{\text{d}}$ are generated from a uniform grid over the feature maps, and the offsets $\Delta p$ are learned through an offset network. All reference points $R$ and offsets $\Delta p$ are in normalized coordinates $[0,1]^2$ per stage, and we clamp $R+\Delta p$ to $[0,1]^2$ before sampling:
\begin{equation}
\Delta p = \theta_{\text{offset}}(Q)
\end{equation}
where $Q$ is the query projected from the input feature map:
\begin{equation}
Q = f W_Q
\end{equation}

The deformed points are then computed as $D_{\text{rgb}} = R_{\text{rgb}} + \Delta p_{\text{rgb}}$ and $D_{\text{d}} = R_{\text{d}} + \Delta p_{\text{d}}$.

At these deformed points, key and value embeddings are sampled using the sampling function $\phi$:
\begin{equation}
\quad \tilde{K} = \tilde{f} W_K, \quad \tilde{V} = \tilde{f} W_V
\end{equation}
where $\tilde{f} = \phi(f; p + \Delta p)$ represents the features sampled at the deformed points. The sampling function $\phi$ is bilinear interpolation, allowing for differentiability:
\begin{equation}
\phi\left(z;(p_x, p_y)\right) = \sum_{(r_x, r_y)} g(p_x, r_x) g(p_y, r_y) z[r_y, r_x, :]
\end{equation}
where $g(a, b) = \max(0, 1 - |a - b|)$. This interpolation ensures that the deformed points smoothly sample from the feature map.

The attention scores are computed using the deformed key embeddings:
\begin{equation}
e_{i,k}^{(m)} = \frac{Q_i^{(m)} \cdot \tilde{K}_k^{(m)\top}}{\sqrt{d_h}}
\end{equation}
where $i$ indexes query positions, $k\!\in\!\{1,\dots,K\}$ indexes the $K$ sampled keys per head, $m$ is the head index, and $d_h=d/M$. We use $K$ sampling locations per head, i.e., $\Delta p\in\mathbb{R}^{N\times M\times K\times 2}$. The model uses $M$ attention heads, where each head performs independent attention computations in parallel.
The final output of an attention head is formulated as:
\begin{equation}
z^{(m)} = \mathrm{Softmax}_k\!\left(e^{(m)} + \phi(\hat{B}; R)\right)\tilde{V}^{(m)}
\end{equation}
where $\hat{B}$ represents the relative position bias and $\phi(\hat{B}; R)$ interpolates the relative position bias table. The outputs of all attention heads are concatenated and projected:
\begin{equation}
f' = \text{Concat}\left(z^{(1)}, \dots, z^{(M)}\right) W_o
\end{equation}
where $W_o$ is the output projection matrix, used to combine the results from all attention heads into a single output representation.
This process results in the refined feature maps $f_{\text{rgb}}'$ and $f_{\text{d}}'$.

\subsection{Deformable Cross-Attention (DC)}
In Deformable Cross-Attention (DC), the model performs cross-modal attention between the features from RGB and depth, using the outputs from the DS module. The inputs to the DC module are the refined features \( f_{\text{rgb}}' \) and \( f_{\text{d}}' \), which are used to generate cross-attended fused features for RGB and depth.

For each modality, the query vectors \( Q_{\text{rgb}}' \) and \( Q_{\text{d}}' \) are obtained from the output of the DS module:
\begin{equation}
Q_{\text{rgb}}' = f_{\text{rgb}}' W_{Q_{\text{rgb}}'}, \quad Q_{\text{d}}' = f_{\text{d}}' W_{Q_{\text{d}}'}
\end{equation}

Next, using an offset network, each query modality predicts offsets that are applied to the reference points of the other modality, allowing us to sample the corresponding key and value embeddings from \( f_{\text{rgb}}' \) or \( f_{\text{d}}' \):
\begin{equation}
\Delta p_{\text{rgb}\rightarrow\text{d}} = \theta_{\text{offset}}(Q_{\text{rgb}}'), \quad
\tilde{f}_{\text{d}}' = \phi(f_{\text{d}}'; R_{\text{d}}' + \Delta p_{\text{rgb}\rightarrow\text{d}})
\end{equation}
\begin{equation}
\Delta p_{\text{d}\rightarrow\text{rgb}} = \theta_{\text{offset}}(Q_{\text{d}}'), \quad
\tilde{f}_{\text{rgb}}' = \phi(f_{\text{rgb}}'; R_{\text{rgb}}' + \Delta p_{\text{d}\rightarrow\text{rgb}})
\end{equation}

From these deformed features, we obtain the key and value embeddings for each modality:
\begin{equation}
\tilde{K}_{\text{rgb}}' = \tilde{f}_{\text{rgb}}' W_{K_{\text{rgb}}'}, \quad \tilde{V}_{\text{rgb}}' = \tilde{f}_{\text{rgb}}' W_{V_{\text{rgb}}'}
\end{equation}
\begin{equation}
\tilde{K}_{\text{d}}' = \tilde{f}_{\text{d}}' W_{K_{\text{d}}'}, \quad \tilde{V}_{\text{d}}' = \tilde{f}_{\text{d}}' W_{V_{\text{d}}'}
\end{equation}

Since this is cross-attention, the query from one modality is used to attend to the keys and values from the other modality. Specifically, the RGB query \( Q_{\text{rgb}}' \) attends to the depth keys and values \( \tilde{K}_{\text{d}}' \) and \( \tilde{V}_{\text{d}}' \), while the depth query \( Q_{\text{d}}' \) attends to the RGB keys and values \( \tilde{K}_{\text{rgb}}' \) and \( \tilde{V}_{\text{rgb}}' \). 

For RGB querying depth and depth querying RGB, the attention scores and output are computed as:
\begin{equation}
z_{\text{rgb}}'^{(m)} = \text{Softmax}\left(Q_{\text{rgb}}'^{(m)} \tilde{K}_{\text{d}}'^{(m)\top} / \sqrt{d_h} + \phi(\hat{B_{\text{rgb}}'}; R_{\text{rgb}}')\right) \tilde{V}_{\text{d}}'^{(m)}
\end{equation}
\begin{equation}
z_{\text{d}}'^{(m)} = \text{Softmax}\left(Q_{\text{d}}'^{(m)} \tilde{K}_{\text{rgb}}'^{(m)\top} / \sqrt{d_h} + \phi(\hat{B_{\text{d}}'}; R_{\text{d}}')\right) \tilde{V}_{\text{rgb}}'^{(m)}
\end{equation}
The outputs from both attention processes, which are the results from the multi-head attention heads, are concatenated to form the final fused feature for the \( i \)-th layer. This fused feature \( f_i \), combining information from both RGB and depth, is passed to the next stage of the model. Here, \( i \) denotes the layer index of the backbone, ensuring that features from each layer are effectively fused to capture complementary information from both modalities.

\subsection{Hierarchical Mask Transformer Decoder}
The fused features from all layers ($f_1, f_2, f_3, f_4$) are passed to the Mask Transformer decoder~\cite{cheng2022masked}, which includes a pixel decoder, Transformer decoder, and mask prediction module. The pixel decoder refines the multi-scale features, while the Transformer decoder predicts mask embeddings. In our class-agnostic approach, the mask prediction module produces a set of mask logits $\hat{M} \in [0, 1]^{Q \times H \times W}$, where each corresponds to a query’s foreground probability map. Binary masks $M \in \{0, 1\}^{Q \times H \times W}$ are obtained by thresholding these logits after upsampling and refinement in the pixel decoder.

\begin{figure}[t!]
\centering
\vspace{2mm}
{\includegraphics[width=0.45\textwidth]{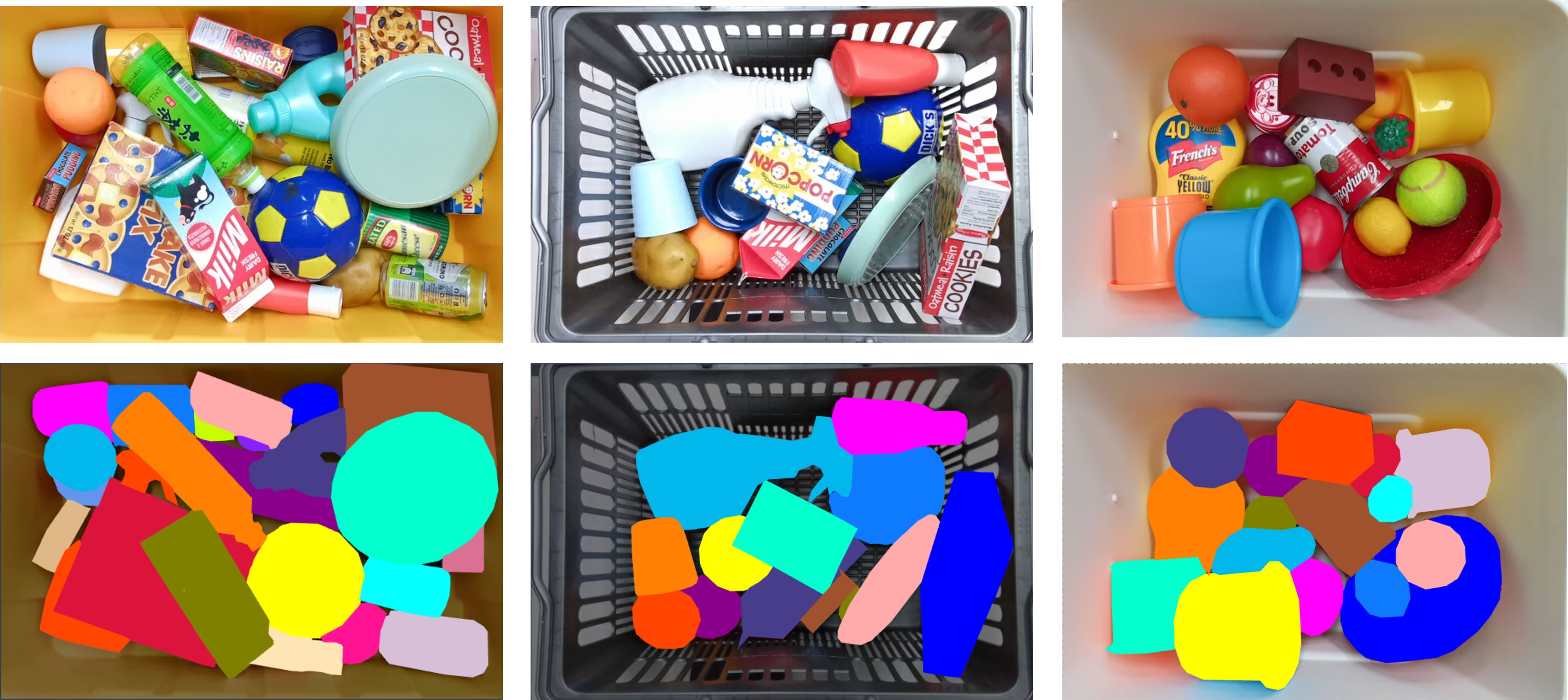}}
\vspace{-2mm}
\caption{\textbf{OCBD Dataset Examples}. Example images from OCBD with their segmentation GTs, illustrating the dataset's diversity and complexity.}
\label{OCBD_dataset}
\vspace{-5mm}
\end{figure}

\begin{table*}[t!]
\vspace{2mm}
\caption{
\centering
{UOIS evaluation results on OCID~\cite{suchi2019easylabel} and OSD~\cite{richtsfeld2012segmentation}.}  }
\label{tab:OCID_OSD}
\centering
\begin{tabular}
{l|c||ccccccc||ccccccc}
\hline
\toprule[1pt]
\multirow{4}{*}{Method} & \multirow{4}{*}{Input} & \multicolumn{7}{c||}{\textbf{OCID (2390 images)~\cite{suchi2019easylabel}}} & \multicolumn{7}{c}{\textbf{OSD (111 images)~\cite{richtsfeld2012segmentation}}} \\
\cline{3-16} & & \multicolumn{3}{c|}{Overlap} & \multicolumn{3}{c|}{Boundary} & \multirow{2}{*}{\%75} & \multicolumn{3}{c|}{Overlap} & \multicolumn{3}{c|}{Boundary} & \multirow{2}{*}{\%75} \\
 & & \multicolumn{1}{c}{P} & \multicolumn{1}{c}{R} & \multicolumn{1}{c|}{F} & \multicolumn{1}{c}{P} & \multicolumn{1}{c}{R} & \multicolumn{1}{c|}{F} &  & \multicolumn{1}{c}{P} & \multicolumn{1}{c}{R} & \multicolumn{1}{c|}{F} & \multicolumn{1}{c}{P} & \multicolumn{1}{c}{R} & \multicolumn{1}{c|}{F} &  \\ 
\bottomrule[1pt]\bottomrule[0pt]

UCN~\cite{xiang2021learning} & RGB & 54.8 & 76.0 & \multicolumn{1}{c|}{59.4} & 34.5 & 45.0 & 36.5 & \multicolumn{1}{|c||}{48.0} & 57.2 & 73.8& \multicolumn{1}{c|}{63.3} & 34.7 & 50.0 & 39.1 & \multicolumn{1}{|c}{52.5} \\

UCN+~\cite{xiang2021learning} & RGB & 59.1 & 74.0 & \multicolumn{1}{c|}{61.1} & 40.8 & 55.0 & 43.8 & \multicolumn{1}{|c||}{58.2} & 59.1 & 71.7 & \multicolumn{1}{c|}{63.8} & 34.3 & 53.3 & 39.5 & \multicolumn{1}{|c}{52.6} \\
           
UOAIS\textsuperscript{†}~\cite{back2022unseen} & RGB & 66.5 & 83.1 & \multicolumn{1}{c|}{67.9} & 62.1 & 70.2 & 62.3 & \multicolumn{1}{|c||}{73.1} & 84.2 & 83.7 & \multicolumn{1}{c|}{83.8} & 72.2 & 72.8 & 72.1 & \multicolumn{1}{|c}{76.7} \\

Mask2Former~\cite{cheng2022masked} & RGB & 67.2 & 73.1 & \multicolumn{1}{c|}{67.1} & 55.9 & 58.1 & 54.5 & \multicolumn{1}{|c||}{54.3} & 60.6 & 60.2 & \multicolumn{1}{c|}{59.5} & 48.2 & 41.7 & 43.3 & \multicolumn{1}{|c}{32.4} \\

MSMFormer~\cite{lu2022mean} & RGB & 70.2 & 84.4 & \multicolumn{1}{c|}{70.5} & 64.5 & 74.9 & 64.9 & \multicolumn{1}{|c||}{75.3} & 59.3 & 82.0 & \multicolumn{1}{c|}{67.9} & 42.9 & 72.0 & 52.4 & \multicolumn{1}{|c}{72.4} \\

MSMFormer+~\cite{lu2022mean} & RGB & 73.9 &  67.1 & \multicolumn{1}{c|}{66.3} & 64.6 & 52.9 & 54.8 & \multicolumn{1}{|c||}{52.8} & 63.9 & 63.7 & \multicolumn{1}{c|}{62.7} & 51.6 & 45.3 & 47.0 & \multicolumn{1}{|c}{41.1} \\

SAM~(ViT-H)\textsuperscript{†}~\cite{kirillov2023segment} & RGB & 8.5 & 64.9 & \multicolumn{1}{c|}{14.0} & 11.1 & 69.5 & 18.3 & \multicolumn{1}{|c||}{45.3} & 9.9 & 60.8 & \multicolumn{1}{c|}{16.3} & 15.3 & 64.3 & 24.4 & \multicolumn{1}{|c}{41.2} \\

\textbf{Ours\textsuperscript{*}} & RGB & 80.4 & 92.6 & \multicolumn{1}{c|}{89.1} & 87.1 & 88.9 & 87.3 & \multicolumn{1}{|c||}{92.6} & 91.5 & 90.7 & \multicolumn{1}{c|}{91.1} & 88.8 & \textbf{88.1} & 88.4 & \multicolumn{1}{|c}{88.6} \\

\bottomrule[0.3pt]
\hline
\toprule[0.3pt]

UCN~\cite{xiang2021learning} & Depth & 83.1 & 90.7 & \multicolumn{1}{c|}{86.4} & 77.7 & 74.3 & 75.6 & \multicolumn{1}{|c||}{75.4} & 78.7 & 83.8 & \multicolumn{1}{c|}{81.0} & 52.6 & 50.0 & 50.9 & \multicolumn{1}{|c}{72.1} \\

UCN+~\cite{xiang2021learning} & Depth & 88.7 & 92.2 & \multicolumn{1}{c|}{90.1} & 83.6 & 84.2 & 83.3 & \multicolumn{1}{|c||}{85.3} & 83.5 & 86.0 & \multicolumn{1}{c|}{84.5} & 57.2 & 56.8 & 56.5 & \multicolumn{1}{|c}{80.6} \\

UOAIS\textsuperscript{†}~\cite{back2022unseen} & Depth & 89.9 & 90.9 & \multicolumn{1}{c|}{89.8} & 86.7 & 84.1 & 84.7 & \multicolumn{1}{|c||}{87.1} & 84.9 & 86.4 & \multicolumn{1}{c|}{85.5} & 68.2 & 66.2 & 66.9 & \multicolumn{1}{|c}{80.8} \\

Mask2Former~\cite{cheng2022masked} & Depth & 82.4 & 79.0 & \multicolumn{1}{c|}{77.4} & 78.2 & 69.1 & 70.4 & \multicolumn{1}{|c||}{66.3} & 79.8 & 81.7 & \multicolumn{1}{c|}{80.4} & 68.4 & 79.0 & 72.2 & \multicolumn{1}{|c}{77.0} \\

\textbf{Ours\textsuperscript{*}} & Depth & 89.0 & 90.3 & \multicolumn{1}{c|}{89.6} & 88.7& 83.6 & 86.1 & \multicolumn{1}{|c||}{83.5} & 87.9 & 86.3& \multicolumn{1}{c|}{87.1} & 73.1 & 67.5 & 70.2 & \multicolumn{1}{|c}{82.9} \\

\bottomrule[0.3pt]
\hline
\toprule[0.3pt]

UCN~\cite{xiang2021learning} & RGB-D & 86.0 & 92.3 & \multicolumn{1}{c|}{88.5} & 80.4 & 78.3 & 78.8 & \multicolumn{1}{|c||}{82.2} & 84.3 & 88.3 & \multicolumn{1}{c|}{86.2} & 67.5 & 67.5 & 67.1 & \multicolumn{1}{|c}{79.3} \\

UCN+~\cite{xiang2021learning} & RGB-D & 91.6 & 92.5 & \multicolumn{1}{c|}{91.6} & 86.5 & 87.1 & 86.1 & \multicolumn{1}{|c||}{89.3} & 87.4 & 87.4 & \multicolumn{1}{c|}{87.4} & 69.1 & 70.8 & 69.4 & \multicolumn{1}{|c}{83.2} \\

UOAIS\textsuperscript{†}~\cite{back2022unseen} & RGB-D & 70.7 & 86.7 & \multicolumn{1}{c|}{71.9} & 68.2 & 78.5 & 68.8 & \multicolumn{1}{|c||}{78.7} & 85.3 & 85.4 & \multicolumn{1}{c|}{85.2} & 72.7 & 74.3 & 73.1 & \multicolumn{1}{|c}{79.1} \\

Mask2Former~\cite{cheng2022masked} & RGB-D & 78.6 & 82.8 & \multicolumn{1}{c|}{79.5} & 69.3 & 76.2 & 71.1 & \multicolumn{1}{|c||}{69.3} & 75.6 & 79.2 & \multicolumn{1}{c|}{77.3} & 54.1 & 64.0 & 58.0 & \multicolumn{1}{|c}{65.2} \\

MSMFormer~\cite{lu2022mean} & RGB-D & 85.3 & 85.4 & \multicolumn{1}{c|}{85.2} & 72.7 & 74.3 & 73.1 & \multicolumn{1}{|c||}{79.1} & 70.7 & 86.7 & \multicolumn{1}{c|}{71.9} & 68.2 & 78.5 & 68.8 & \multicolumn{1}{|c}{78.7} \\

MSMFormer+~\cite{lu2022mean} & RGB-D & 92.5 & 91.0 & \multicolumn{1}{c|}{91.5} & 89.4 & 85.9 & 87.3 & \multicolumn{1}{|c||}{86.0} & 87.1 & 86.1 & \multicolumn{1}{c|}{86.4} & 69.0 & 68.6 & 68.4 & \multicolumn{1}{|c}{80.4} \\

\textbf{Ours~(DS)} & RGB-D & 92.8 & 92.3 & \multicolumn{1}{c|}{91.8} & 91.3 & 89.4 & 89.7 & \multicolumn{1}{|c||}{92.7} & 93.0 & 92.1& \multicolumn{1}{c|}{92.5} & 91.2 & 85.4 & \textbf{88.2} & \multicolumn{1}{|c}{92.8} \\

\textbf{Ours~(DS+DC)} & RGB-D & \textbf{93.2} & \textbf{92.6} & \multicolumn{1}{c|}{\textbf{92.1}} & \textbf{91.8}& \textbf{89.8} & \textbf{90.0} & \multicolumn{1}{|c||}{\textbf{92.9}} & \textbf{93.5} & \textbf{92.4}& \multicolumn{1}{c|}{\textbf{92.9}} & \textbf{91.7} & 84.6 & 88.0 & \multicolumn{1}{|c}{\textbf{92.9}} \\

\bottomrule[1pt] \hline
\end{tabular}
\begin{tablenotes}       
    \item[1]\ + : The method using second-stage network (Zoom-in refinement).
    \item[2]\ \textsuperscript{†} : The model that incorporate a foreground segmentation model~\cite{wu2020cgnet} in OSD.
    \item[3]\ * : The model after passing input through the DS module without fusion.
\end{tablenotes}
\end{table*}

\section{Object Clutter Bin Dataset (OCBD)}
Existing UOIS benchmarks~\cite{richtsfeld2012segmentation, suchi2019easylabel} primarily focus on front-view scenarios with stacked or overlapping objects, which are insufficient for evaluating logistics tasks like bin-picking, where objects may be randomly piled in multiple layers from a top-down view. These conditions present challenges for segmentation models, requiring them to handle cluttered scenes with diverse shapes, orientations, and occlusions. Earlier bin-picking datasets~\cite{xu2022fpcc, xu2020convolutional} predominantly feature industrial parts—repetitive, texture-less objects—and are often simulated~\cite{chen2021metagraspnet_v0, gilles2022metagraspnet}, lacking the noise and variability of real-world scenarios.

To address these limitations, we introduce the Object Clutter Bin Dataset (OCBD), reflecting real-world bin environments. OCBD consists of 1000 top-down RGB-D images (640×480) captured with an Azure Kinect DK. All ground truth segmentation labels were human-annotated for high-quality instance masks even in challenging cluttered scenes. The dataset features 30 YCB objects~\cite{calli2017yale} and 25 HOPE household items~\cite{tyree20226}, offering diversity in shape, size, and texture, and includes various bins with different backgrounds and lighting to enhance realism.

With 13,147 object instances and an average of 13.2 objects per image, OCBD offers a challenging benchmark. Objects often overlap or block one another, contrasting with datasets like OSD~\cite{richtsfeld2012segmentation} and OCID~\cite{suchi2019easylabel}, where fewer objects appear per scene (7.5 and 3.3, respectively) in more organized setups. By focusing on realistic cluttered bin-picking scenarios, OCBD provides a critical resource for developing and evaluating segmentation models in real-world robotic manipulation.

\section{EXPERIMENTS}

\subsection{Implementation Details}
\textbf{Dataset and Preprocessing.} 
We train the model on the UOAIS-SIM dataset~\cite{back2022unseen}, which contains 50,000 photorealistic RGB-D images across diverse logistics scenarios, including both tabletop and bin environments. The depth input is converted into HHA format~\cite{gupta2014learning}, where the three channels represent disparity, height above ground, and the angle between the local surface normal and the gravity direction. During training, we apply standard data augmentation techniques such as random cropping, horizontal flipping, and scaling to increase the diversity of training samples.

\textbf{Training Setup.} 
We train the model for 20 epochs using the AdamW optimizer~\cite{loshchilov2017decoupled} with a learning rate of $1e^{-4}$ and a batch size of 2. A cosine learning rate scheduler with warm-up is employed to improve convergence. All experiments are conducted on an NVIDIA A100 GPU, with a memory allocation of 40GB.

\textbf{Loss Functions.} We utilize the loss function used in Mask2Former~\cite{cheng2022masked}, adapted for UOIS. The mask loss combines binary cross-entropy and dice loss~\cite{milletari2016v}, expressed as \( L_{\text{mask}} = \lambda_{\text{ce}} L_{\text{ce}} + \lambda_{\text{dice}} L_{\text{dice}} \), where both \( \lambda_{\text{ce}} \) and \( \lambda_{\text{dice}} \) are set to 5.0. The final loss is computed as the sum of the mask loss and a classification loss that distinguishes between foreground (object) and background (non-object): \( L_{\text{final}} = L_{\text{mask}} + \lambda_{\text{cls}} L_{\text{cls}} \), where \( \lambda_{\text{cls}} \) is 2.0 for predictions matching a foreground object and 0.1 for predictions classified as background.

\begin{table}[t!]
\centering
\caption{ \centering
UOIS Evaluation Results on OCBD with RGB-D Input.}
\label{tab:seg_OCBD}
\resizebox{0.95\columnwidth}{!}{%
\renewcommand{\arraystretch}{1.3}
\begin{tabular}
{l||ccccccc}
\toprule[1.2pt]
\multirow{2}{*}{Method} & \multicolumn{7}{c}{\textbf{OCBD (1000 images)}}\\

\cline{2-8} & \multicolumn{3}{c|}{Overlap} & \multicolumn{3}{c|}{Boundary} & \multirow{2}{*}{\%75} \\
 & \multicolumn{1}{c}{P} & \multicolumn{1}{c}{R} & \multicolumn{1}{c|}{F} & \multicolumn{1}{c}{P} & \multicolumn{1}{c}{R} & \multicolumn{1}{c|}{F} &  \\ 
\bottomrule[1pt]\bottomrule[0pt]


UCN+~\cite{xiang2021learning} & 76.7 &  73.2 & \multicolumn{1}{c|}{70.7} & 50.7 &  51.9 & \multicolumn{1}{c|}{46.7} & 61.3 \\ 

UOAIS~\cite{back2022unseen} & 68.3 & 77.9 & \multicolumn{1}{c|}{69.9} & 55.7 & 57.3 & \multicolumn{1}{c|}{54.9} & 58.2 \\ 


MSMFormer+~\cite{lu2022mean} & 86.6 & 34.9 & \multicolumn{1}{c|}{42.5} & 66.2 &  23.4 & \multicolumn{1}{c|}{28.6} & 26.5 \\ 

\textbf{Ours~(DS)} & 91.3 & 90.1 &  \multicolumn{1}{c|}{90.4} & 89.4 & 86.2 & \multicolumn{1}{c|}{87.3} & 85.8 \\ 

\textbf{Ours~(DS+DC)} & \textbf{92.5} & \textbf{90.4} &  \multicolumn{1}{c|}{\textbf{91.3}} & \textbf{90.8} & \textbf{87.3} & \multicolumn{1}{c|}{\textbf{88.7}} & \textbf{87.1} \\ 

\bottomrule[1.2pt]
\end{tabular}%
}
\vspace{-3mm}
\end{table}

\begin{figure*} [t!]
\centering
\vspace{2mm}
\includegraphics[width=0.95\textwidth]{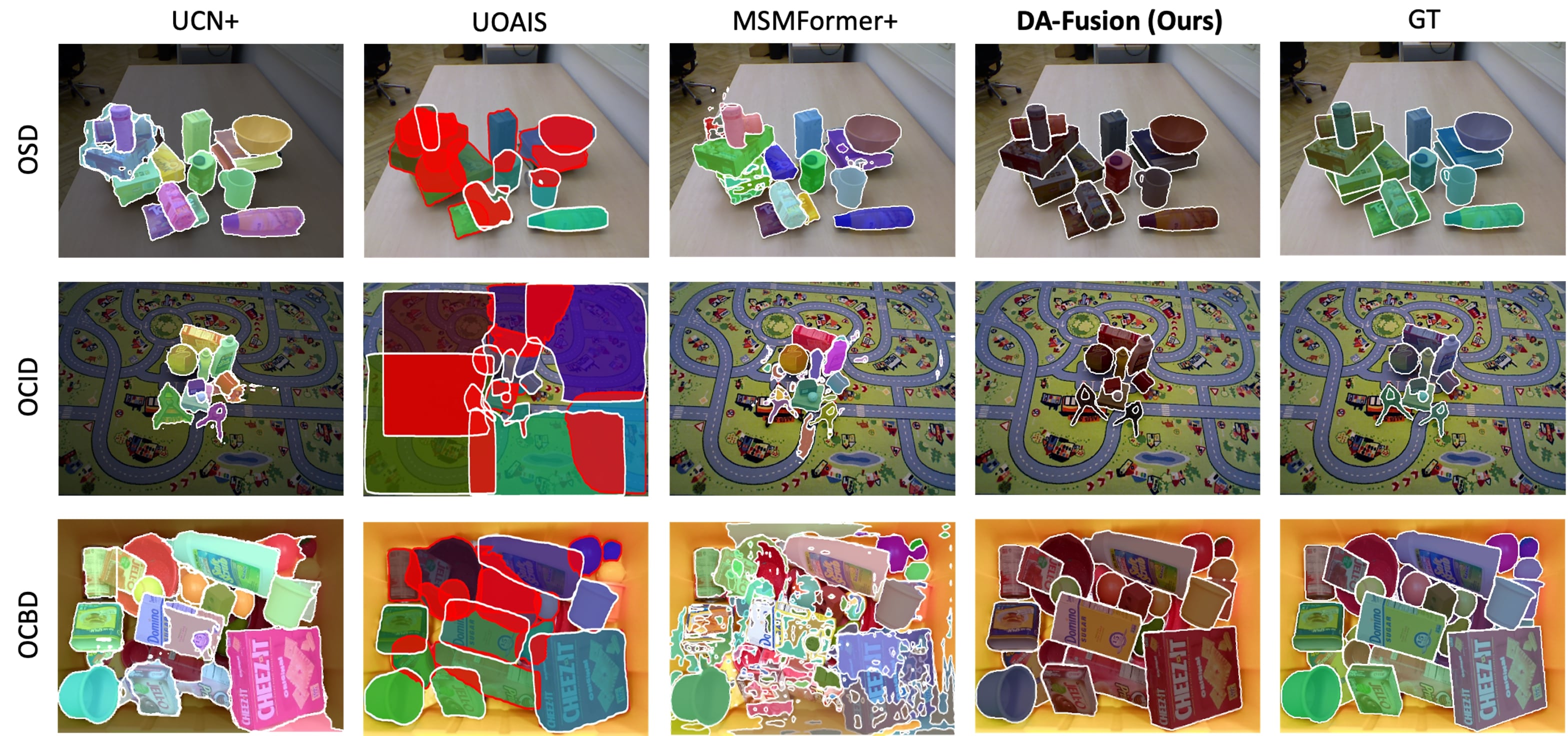}
\caption{\textbf{Qualitative Evaluation.} Comparison of instance segmentation results across OSD (top row), OCID (middle row), and OCBD (bottom row) datasets. From left to right: UCN, UOAIS, MSMFormer, our DA-Fusion method, and GT. DA-Fusion demonstrates superior segmentation of accumulated objects in OSD, better object delineation in feature-rich backgrounds in OCID, and more accurate detection of occluded objects in lower layers in OCBD, as evidenced by the closer resemblance to GT.}
\label{result}
\vspace{-2mm}
\end{figure*}

\subsection{Experimental Setup}
\textbf{Datasets.} 
We evaluated DA-Fusion on the UOIS task using three distinct benchmarks: OSD~\cite{richtsfeld2012segmentation} (111 images), OCID~\cite{suchi2019easylabel} (2,390 images), and OCBD (1,000 images), our newly introduced dataset designed to reflect real-world bin-picking tasks with cluttered, multi-layered object arrangements from a top-down view. These datasets cover a range of logistics scenarios, from complex tabletop settings to cluttered bin-picking operations in warehouses.

\textbf{Evaluation Metrics.}  
Object segmentation performance is evaluated using Overlap and Boundary Precision (P), Recall (R), and F-measure (F). These are calculated as \( P = \frac{TP}{TP + FP} \), \( R = \frac{TP}{TP + FN} \), and \( F = \frac{2PR}{P + R} \), where \( TP \), \( FP \), and \( FN \) denote true positives, false positives, and false negatives. We also report the percentage of segmented objects with an Overlap F-measure \( \geq 75\% \) to evaluate performance in cluttered environments.

\subsection{Quantitative Results}
As shown in Table~\ref{tab:OCID_OSD}, our RGB-D fusion method achieved the highest overall scores on the OCID and OSD datasets, significantly outperforming previous state-of-the-art approaches. Prior methods typically relied on two-stage clustering models~\cite{xiang2021learning, lu2022mean} to refine segmentation masks in cluttered scenes, whereas our approach maintained superior accuracy without additional computational overhead.
 
The results on the OCBD dataset, shown in Table~\ref{tab:seg_OCBD}, further demonstrate the effectiveness of DA-Fusion in complex bin-picking scenarios. Compared to the same set of methods, our model consistently outperforms across both tabletop and bin-picking tasks, showing robust performance when handling unseen objects in cluttered, dynamic environments—crucial for real-world logistics applications.

\subsection{Qualitative Results}
Fig.~\ref{result} presents qualitative comparisons, where DA-Fusion consistently demonstrates superior segmentation in complex scenes with diverse object shapes and occlusions, whereas other methods struggle in such environments. Specifically, in the OSD dataset, other methods tend to fail at separating accumulated objects, while our method accurately distinguishes individual items. For the OCID dataset, which contains backgrounds with rich features, our method significantly outperforms alternatives in properly delineating objects. Finally, in the OCBD dataset representing cluttered bin scenarios, our approach effectively segments objects even in lower layers, where other methods like UOAIS either miss objects entirely or produce incomplete segmentations.

\begin{figure} [t!]
\centering
{\includegraphics[width=0.75\columnwidth]{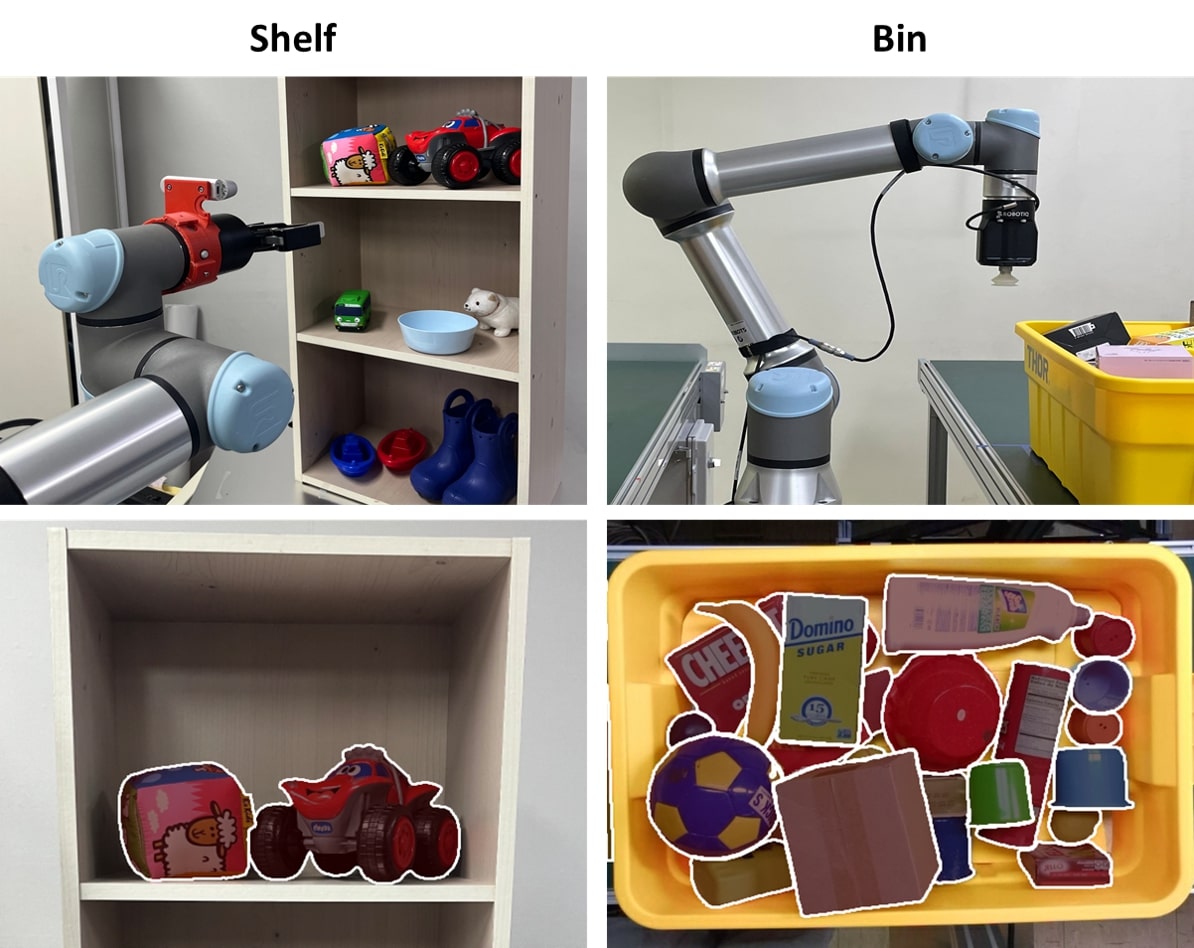}}
\caption{\textbf{Examples from Real-Robot Experiments.} Shelf picking (left column) and bin picking (right column) tasks with corresponding segmentation results demonstrating effectiveness of DA-Fusion in diverse real-world manipulation environments.}
\label{setting_fig}
\vspace{-5mm}
\end{figure}

\subsection{Real-robot Experiments}
We validated the effectiveness of DA-Fusion through real-robot experiments focused on two logistics tasks: shelf-picking from a front-view setup and bin-picking from a top-down view. Fig.~\ref{setting_fig} shows these experiments, demonstrating how the model adapts to different real-world conditions.
In the shelf-picking task, we used a UR5e robotic arm equipped with a two-finger gripper and a RealSense D435 camera mounted above the gripper for RGB-D data capture. For bin-picking, we used an Azure Kinect DK camera positioned overhead.
The accurate segmentation of unseen objects improved the success rate of robotic manipulation across both tasks. These results highlight the method's robustness in handling cluttered and dynamic environments, showing its potential for improving automation in logistics operations.

\section{CONCLUSION}
We introduce DA-Fusion, a deformable attention-based RGB-D fusion model for unseen object instance segmentation in logistics tasks like shelf (front-view) and bin picking (top-down view). Our approach significantly improves segmentation accuracy in challenging environments. Additionally, we present OCBD, a new benchmark for realistic bin-picking scenarios. Experimental results, including real-robot tests, demonstrate DA-Fusion's superior performance and adaptability across various logistics applications.







\bibliographystyle{ieeetr}
\bibliography{ref}

\end{document}